\documentclass[11pt]{article}

\usepackage[english]{babel}
\usepackage[a4paper,top=2cm,bottom=2cm,left=3cm,right=3cm]{geometry}

\usepackage{setspace}
\usepackage{rotating}

\usepackage{amsmath,amssymb,amsfonts}
\usepackage{amsthm}
\usepackage{mathrsfs}

\usepackage{graphicx}
\usepackage{booktabs}
\usepackage{multirow}
\usepackage{adjustbox}
\usepackage{longtable}
\usepackage{float}
\usepackage{subcaption}

\usepackage{caption}
\usepackage{etoolbox}

\AtBeginEnvironment{abstract}{\begingroup\singlespacing}
\AtEndEnvironment{abstract}{\endgroup}

\newcommand{\keywordsinabstract}[1]{%
  \vspace{0.2cm}%
  {\footnotesize \emph{Keywords}: #1\par}%
  \vspace{0.2cm}%
}

\usepackage[numbers]{natbib}
\usepackage[colorlinks=true,allcolors=blue]{hyperref}

\usepackage{enumitem} 

\usepackage[ruled,vlined]{algorithm2e}
\SetKwComment{Comment}{/* }{ */}

\usepackage{authblk}

\usepackage[hang]{footmisc}

\usepackage{xcolor}
\usepackage{placeins}
\usepackage[title]{appendix}
\usepackage{textcomp}
\usepackage{manyfoot}
\usepackage{listings}

\title{Uncertainty-Aware Delivery Delay Duration Prediction via Multi-Task Deep Learning}

\author[1]{Stefan Faulkner}
\author[1]{Reza Zandehshahvar}
\author[2]{Vahid Eghbal Akhlaghi}
\author[2]{Sebastien Ouellet}
\author[2]{Carsten Jordan}
\author[1]{Pascal Van Hentenryck}

\affil[1]{NSF Artificial Intelligence Institute for Advances in Optimization, \newline 
H. Milton Stewart School of Industrial and Systems Engineering, \newline 
Georgia Institute of Technology, Atlanta, GA, USA}
\affil[2]{Kinaxis}

\affil[ ]{\texttt{\{sfaulkner9, mohammadreza3, pvh\}@gatech.edu}}
\affil[ ]{\texttt{\{vakhlaghi, souellet, cjordan\}@kinaxis.com}}

\begin{document}

\date{}
\maketitle

\begin{abstract}
Accurate delivery delay prediction is critical for maintaining operational efficiency and customer satisfaction across modern supply chains. Yet the increasing complexity of logistics networks, spanning multimodal transportation, cross-country routing, and pronounced regional variability, makes this prediction task inherently challenging. This paper introduces a multi-task deep learning model for delivery delay duration prediction in the presence of significant imbalanced data, where delayed shipments are rare but operationally consequential. The model embeds high-dimensional shipment features with dedicated embedding layers for tabular data, and then uses a classification-then-regression strategy to predict the delivery delay duration for on-time and delayed shipments. Unlike sequential pipelines, this approach enables end-to-end training, improves the detection of delayed cases, and supports probabilistic forecasting for uncertainty-aware decision making. The proposed approach is evaluated on a large-scale real-world dataset from an industrial partner, comprising more than 10 million historical shipment records across four major source locations with distinct regional characteristics. The proposed model is compared with traditional machine learning methods. Experimental results show that the proposed method achieves a mean absolute error of 0.67--0.91 days for delayed-shipment predictions, outperforming single-step tree-based regression baselines by 41--64\% and two-step classify-then-regress tree-based models by 15--35\%. These gains demonstrate the effectiveness of the proposed model in operational delivery delay forecasting under highly imbalanced and heterogeneous conditions.

\keywordsinabstract{Delivery Delay Prediction, Multitask Learning, Supply Chains, Deep Learning, Quantile Regression, Uncertainty Quantification, Conformal Prediction}
\end{abstract}

\newpage 

\section{Introduction}

Accurate delivery delay prediction has become a critical requirement in modern supply chains. Manufacturers, retailers, and logistics service providers increasingly rely on reliable delivery estimates to plan inventory, allocate resources, honor service level agreements, and maintain customer satisfaction~\cite{vanderspoelPredictiveAnalyticsTruck2016, liuPredictingPurchaseOrders2018}. As logistic networks grow in scale and complexity, and with the rapid expansion of e-commerce, the uncertainty in delivery times increases. This can lead to cascading operational and financial consequences, including stockouts, misallocated labor, and eroded service quality~\cite{balsterETAPredictionModel2020, mohammadInnovativeSolutionsLast2023}. Early and accurate identification of potential delays is therefore critical for initiating contingency plans, optimizing resource deployments, and providing trustworthy customer communication~\cite{fisherValueRapidDelivery2019, masuchFasterBetterImpact2024}.

Traditional approaches for delivery time prediction rely on rule based or static service level driven estimation methods. However, these methods fail to adapt to the dynamic changes in modern supply chains caused by factors such as demand volatility and localized operational conditions that govern real world delivery performance~\cite{urciuoliAlgorithmImprovedETAs2018, gabelliniDeepLearningApproach2024}. As a result, the logistics industry is increasingly turning toward data-driven machine learning (ML) methods. Historical shipment data contains fine-grained signals reflecting seasonal patterns, regional behavior, transportation modes, and interactions among a large number of shipment attributes, which could be utilized for data-driven prediction of the delivery time and delays.

Despite the growing adoption of machine learning methods for delivery delay prediction, the existing literature exhibits several important limitations. First, as highlighted by a recent systematic review from Muller et al.~\cite{muller_analytical_2025}, existing research is heavily skewed toward binary classification (predicting if a shipment is late) rather than regression (predicting the duration of delay). Second, uncertainty-aware estimation remains underexplored, despite its importance for downstream decision making under operational risk. Third, the significant class imbalance between on-time and delayed shipments requires approaches that can handle class imbalance and provide reliable predictions for both majority and minority classes.

This paper introduces a multi-task deep learning model for delivery delay prediction that jointly addresses point estimation and uncertainty quantification. The paper presents a classification-then-regression model within a single end-to-end architecture, enabling effective learning under severe class imbalance and reducing the error propagation commonly observed in sequentially trained pipelines. In addition, the model produces calibrated prediction intervals, allowing uncertainty-aware decision making in operational settings.

The main contributions of this work are summarized as follows:
\begin{itemize}
    \item A unified multi-task learning model for uncertainty-aware delivery delay prediction. The proposed model utilizes dedicated embedding layers for categorical and numerical features to handle heterogeneous and high-dimensional tabular data of the shipments, and integrates conformal prediction to improve the probabilistic forecasts.

    \item A comprehensive empirical evaluation on a large-scale, real-world shipment dataset with over 10 million instances, demonstrating consistent improvements over widely used tree-based baselines, including XGBoost and CatBoost. The dataset is provided by an industrial partner, and includes shipment records from four source locations to more than 190k destinations from 2022 to 2024.
\end{itemize}

Experimental results show that the proposed method achieves a mean absolute error (MAE) of 0.67–0.91 days for delayed-shipment predictions, outperforming single-step tree-based regression models by 41–64\% and two-step classify-then-regress tree-based approaches by 15–35\%. Moreover, the proposed model yields more reliable probabilistic forecasts, achieving higher empirical coverage with sharper prediction intervals.

The remainder of this paper is organized as follows. Section 2 reviews related work. Sections 3 and 4 present the problem statement and the methodology. Section 5 describes the experimental setup and dataset. The experimental results are presented in Section 6, followed by the conclusion in Section 7.


\section{Related Work}

\subsection{Delivery Systems and Delay}

Timely and reliable delivery has become a critical driver of performance in modern supply chains and can directly impact sales, profitability, and customer satisfaction. Fisher et al.~\cite{fisherValueRapidDelivery2019} reported that reducing delivery times can increase sales by up to 4\% and profits by 2.2\%. However, subsequent studies revealed a more nuanced relationship: while shorter promised delivery times can boost sales, overly aggressive commitments often lead to delays, increased return rates, and weaker customer retention \cite{cuiSoonerLaterPromising2024, masuchFasterBetterImpact2024}. These findings emphasize that both the speed and reliability of delivery critically shape customer behavior and downstream business outcomes. The impact of delivery time uncertainty on order fulfillment and carrier selection is further studied by Ye et al.~\cite{yeContextualStochasticOptimization2024b}, emphasizing the importance of reliable delivery time estimates for optimizing operational costs.

Uncertainty in delivery times arises from a wide range of operational factors, particularly in supply chains that rely on multimodal transportation. Prior work has examined delays in maritime shipping, showing that weather conditions, port congestion, and routing decisions contribute significantly to vessel arrival variability \cite{kimEarlyDetectionVessel2017}. Other studies investigated the cascading effect of delays at various stages of transportation and how disruptions at one step can amplify system-wide delivery time variability ~\cite{urciuoliAlgorithmImprovedETAs2018, baryannisPredictingSupplyChain2019}. Similar complexities appear in last-mile delivery, where urban congestion, service type, and carrier selection create significant fluctuations that cannot be explained by static planning assumptions {\cite{slabinacINNOVATIVESOLUTIONSLASTMILE2015, mohammadInnovativeSolutionsLast2023}}. These studies collectively illustrate that delays are rarely isolated events; rather, they arise from interconnected and context-dependent sources across the logistics network.

The diverse and interdependent causes of delivery-time variability, spanning omnichannel fulfillment, maritime transport, and last-mile operations, underscore the need for more sophisticated prediction methods. Traditional rule-based approaches, which assume fixed transit times or rely on static service levels, fail to capture the dynamic and nonlinear interactions that shape real-world delivery performance. This gap has motivated the development of data-driven ML models capable of learning complex delay patterns from historical operational data and providing more accurate, reliable delivery time predictions.

\subsection{Machine Learning for Delivery Time and Delay Prediction}
Machine learning approaches have been increasingly applied to delivery-time estimation and delay prediction in logistics and supply chain networks. While many studies focus on predicting the absolute delivery time of a shipment \cite{vanderspoelPredictiveAnalyticsTruck2016, khiariBoostingAlgorithmsDelivery2020a, wuDeepETASpatialTemporalSequential2019, rokossCaseStudyDelivery2024}, this work centers on delay prediction, which seeks to estimate the deviation between the actual arrival time and the promised delivery time.

Recent research has formulated delay prediction as both a classification and a regression problem. In the classification setting, the objective is to determine whether a shipment will arrive on time or be delayed. Ensemble learning techniques have been applied to classify delayed shipments using historical shipment records and operational features (e.g., origin–destination characteristics, service type, and carrier information) {\cite{abouloifaPredictingLateDelivery2022, rezki2024machine}}. Deep learning models have also been evaluated against traditional tree-based approaches, demonstrating improved performance in capturing complex feature interactions {\cite{bassiouni_deep_2024}}. In the regression setting, other studies have proposed gradient boosting or neural network–based models to estimate the magnitude of delay directly {\cite{steinbergNovelMachineLearning2023a}}, including methods that incorporate external signals such as macroeconomic indicators to improve predictive accuracy {\cite{gabelliniDeepLearningApproach2024}}.

Despite these advances, several important limitations remain, as highlighted in the systematic review by Müller et al.\ {\cite{muller_analytical_2025}}. Most existing studies rely on datasets where the proportion of delayed shipments is relatively balanced {\cite{rezki2024machine, abouloifaPredictingLateDelivery2022, bassiouni_deep_2024}}. However, in real-world operations, the vast majority of shipments arrive on time, leading to severe class imbalance that can significantly degrade model performance. Moreover, uncertainty quantification, including prediction intervals for delay duration, has received limited attention, even though accurate assessments of uncertainty are essential for operational decision-making. These gaps underscore the need for delay-prediction methods that can robustly handle imbalanced data and offer calibrated, uncertainty-aware estimates.

In addition, the shipment delivery data often includes heterogeneous tabular data, which introduces challenges for vanilla neural networks. The recent advances in deep learning for tabular data have introduced embedding techniques that improve model performance on heterogeneous features~\cite{robinson2024relbench, borisov2022deep, arik2021tabnet}. Gorishniy et al.~\cite{gorishniyEmbeddingsNumericalFeatures2023a} demonstrated that piece-wise linear encoding and quantile binning can enhance deep learning performance on tabular data. These embedding approaches have not yet been fully explored in the context of delivery delay duration estimation.

\subsection{Multi-task Learning  and Uncertainty Quantification}

Multi-task learning (MTL) helps in improving the performance of interrelated tasks by leveraging shared representations, as introduced by Caruana~\cite{caruana_multitask_1997}. In deep neural networks, MTL is typically implemented through hard parameter sharing, where hidden layers are shared across tasks while maintaining task-specific output layers. MTL has been increasingly applied in transportation and logistics: Tang et al.~\cite{tang_adaptive_2024} developed an adaptive multi-task model for route-specific travel time estimation, Roy et al.~\cite{roy_multi-task_2025} jointly predicted traffic emissions and travel delay, and Khalil and Fatmi~\cite{khalil_how_2025} demonstrated that hard parameter sharing improves performance when jointly modeling classification and regression tasks. For delivery time estimation, Zhang et al.~\cite{zhangDualGraphMultitask2023} proposed a dual graph multi-task framework addressing imbalanced data by separately modeling frequent and rare delivery patterns, while Yi et al.~\cite{yi_learning_2024} developed TransPDT, a Transformer-based model deployed at JD Logistics. Despite these advances, the application of MTL under severe class imbalance remains underexplored in the context of delivery delay prediction.

Beyond point predictions, uncertainty quantification is essential for operational decision-making. Quantile regression~\cite{Koenker_2005} estimates conditional quantiles rather than mean values, providing richer distributional information. In delivery time prediction, Zhang et al.~\cite{zhangDeliveryTimePrediction2023} and Rueda-Toicen and Zea~\cite{rueda-toicenEstimatingParcelDelivery2024} applied quantile regression to obtain prediction intervals. However, quantile regression often produces intervals with insufficient coverage during inference. Conformal prediction addresses this limitation by providing distribution-free coverage guarantees~\cite{conformalvovk, shaferTutorialConformalPrediction2007}. Romano et al.~\cite{romanoConformalizedQuantileRegression2019} proposed Conformalized Quantile Regression (CQR), which combines quantile regression's ability to capture heteroscedasticity with conformal prediction's calibration properties. Within logistics, Liu et al.~\cite{liu_uncertainty-aware_2023} developed ProbTTE at DiDi, demonstrating the operational value of probabilistic forecasting at industrial scale, while Ye et al.~\cite{ye_conformal_2025} applied conformal methods to order fulfillment forecasting. However, these approaches focus on delivery time estimation rather than delay prediction, and none addresses the zero-inflated distributions characteristic of delay data where most shipments arrive on time. This paper combines MTL with conformalized quantile regression, providing reliable prediction intervals while handling the zero-inflated structure through specialized dual-head routing.

\begin{table}[t!]
\caption{Mathematical notations and definitions.}
\label{tab:notation}
\resizebox{\textwidth}{!}{%
\begin{tabular}{lp{4.5cm}p{8.5cm}}
\toprule
\textbf{Symbol} & \textbf{Domain/Type} & \textbf{Description} \\
\midrule
\multicolumn{3}{l}{\textit{Indices and Counts}} \\
\midrule
$N$ & Positive integer & Number of shipments \\
$K$ & Positive integer & Number of categorical features \\
$M$ & Positive integer & Number of numerical features \\
$i$ & $\{1,\ldots,N\}$ & Shipment index \\
$k$ & $\{1,\ldots,K\}$ & Categorical feature index \\
$m$ & $\{1,\ldots,M\}$ & Numerical feature index \\
\midrule
\multicolumn{3}{l}{\textit{Data and Features}} \\
\midrule
$\mathcal{D}_{\text{train}}$, $\mathcal{D}_{\text{val}}$, $\mathcal{D}_{\text{calib}}$, $\mathcal{D}_{\text{test}}$ & & Training, validation, calibration, and test datasets \\
$\mathbf{x}_i$ & & Feature vector for shipment $i$ ($K$ categorical + $M$ numerical) \\
$\mathbf{x}_i^{\text{cat}}$, $\mathbf{x}_i^{\text{num}}$ & & Categorical and numerical feature subvectors \\
$\mathbf{z}_i$ & $\mathbb{R}^{d_z}$ & Concatenated embedded feature vector \\
$y_i$ & $\mathbb{R}$ & Delay duration in days (negative = early, positive = delayed) \\
$d_i$ & $\{0,1\}$ & Binary delay indicator ($d_i = 1$ if $y_i \geq 1$) \\
\midrule
\multicolumn{3}{l}{\textit{Model Components}} \\
\midrule
$\mathbf{h}_i$ & $\mathbb{R}^{d_h}$ & Learned representation from shared MLP backbone \\
$f_c$ & & Classification head \\
$f_r^{\text{delayed}}$, $f_r^{\text{ontime}}$ & & Dual regression heads for delayed and on-time predictions \\
$\hat{p}_i$ & $(0,1)$ & Predicted delay probability for shipment $i$ \\
$\hat{d}_i$ & $\{0,1\}$ & Predicted binary delay indicator \\
$\mathbf{r}_i$ & $\mathbb{R}^{d_h+1}$ & Regression input vector $[\mathbf{h}_i, \hat{p}_i]$ \\
$\hat{y}_i^{(\alpha)}$ & $\mathbb{R}$ & Predicted $\alpha$-quantile of delay duration \\
\midrule
\multicolumn{3}{l}{\textit{Loss Functions}} \\
\midrule
$\mathcal{L}_{t}$ & & Total loss: $\mathcal{L}_{c} + \mathcal{L}_{r}$ \\
$\mathcal{L}_{c}$ & & Classification loss (SigmoidF1) \\
$\mathcal{L}_{r}$ & & Regression loss (pinball loss) \\
$\rho_{\alpha}(y_i, \hat{y}_i^{(\alpha)})$ & & Pinball loss for quantile level $\alpha$ \\ 
\midrule
\multicolumn{3}{l}{\textit{Uncertainty Quantification}} \\
\midrule
$\alpha$ & $(0,1)$ & Miscoverage level (e.g., $\alpha = 0.2$ for 80\% coverage) \\
$\hat{C}^{\alpha}(\mathbf{x}_i)$ & & Initial prediction interval $[\hat{y}_i^{(\alpha/2)}, \hat{y}_i^{(1-\alpha/2)}]$ \\
$E_i$ & $\mathbb{R}$ & Conformity score for sample $i$ in calibration \\
$\hat{q}_{1-\alpha}$ & $\mathbb{R}$ & Conformal correction term ($(1-\alpha)$-quantile of $\{E_i\}$) \\
$\tilde{C}^{\alpha}(\mathbf{x}_i)$ & & Calibrated prediction interval \\
\bottomrule
\end{tabular}
}
\end{table}

\section{Problem Statement}

\begin{figure}[t!]
    \centering
    \includegraphics[width=0.8\textwidth]{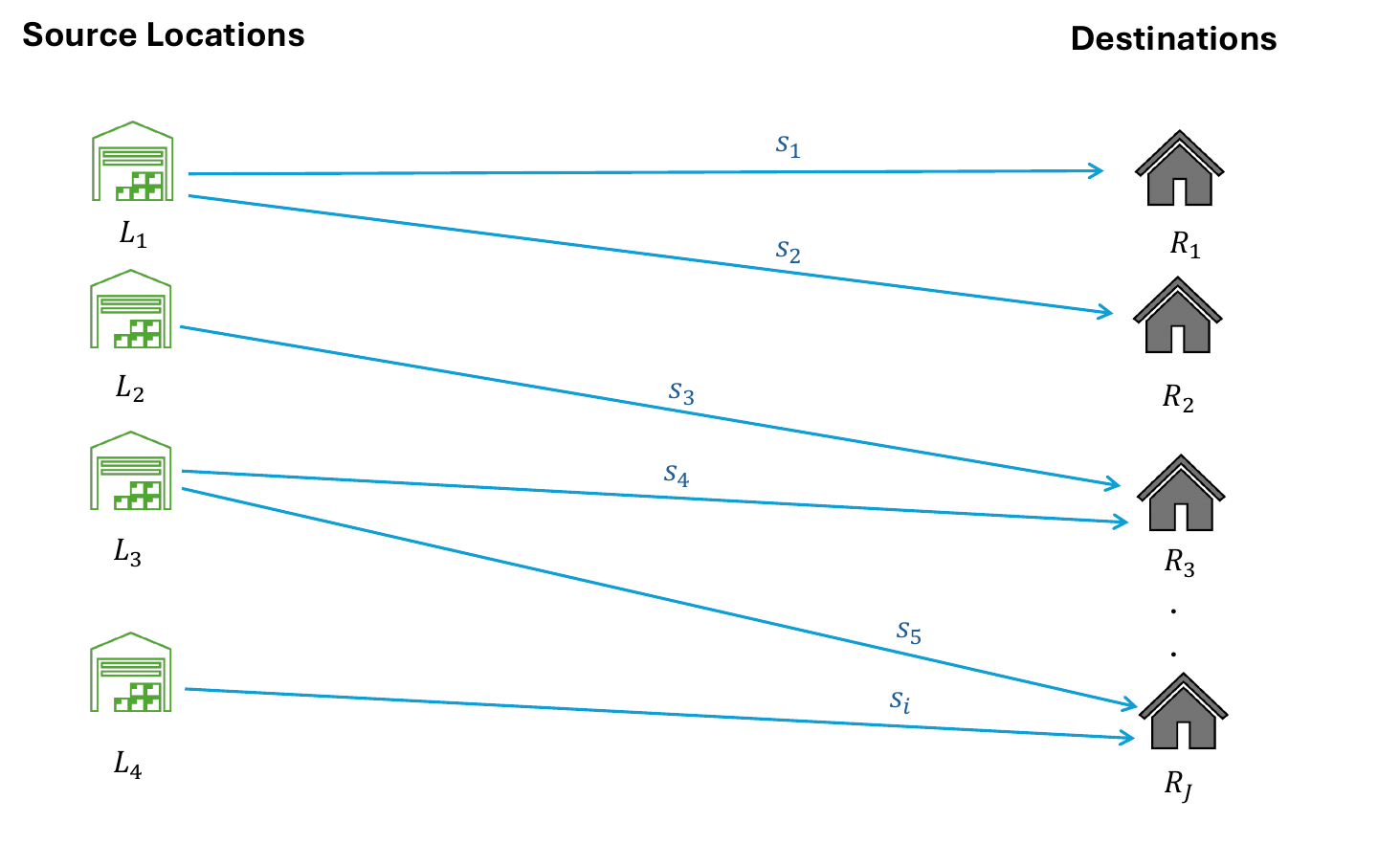}
    \caption{Logistics network overview. Shipments $s_i \in \mathcal{S}$ flow from origin locations $L_k \in \mathcal{L}$ to destinations $R_j \in \mathcal{R}$, characterized by physical features (weight, volume, items count) and timing information (planned departure/arrival).}
    \label{fig:logistics-network}
\end{figure}

Consider a logistics network with a set of source locations $\mathcal{L}$ and a set of destination locations $\mathcal{R}$, as illustrated in Figure~\ref{fig:logistics-network}. Table~\ref{tab:notation} summarizes the mathematical notation used throughout this paper.

Each shipment $s_i \in \mathcal{S}$, for $i \in \{1, \dots, N\}$, is described by a feature vector $\mathbf{x}_i \in \mathcal{X}$ that contains its origin--destination pair, physical attributes (e.g., weight, volume), and service-level characteristics such as carrier, service type, and handling instructions. A detailed description of the feature set is provided in Section~\ref{sec: Data}.

For each shipment $s_i$, let $t_i^{\text{planned}}$ denote the planned delivery time determined by the enterprise rule-based system at the time of order placement, and let $t_i^{\text{actual}}$ denote the actual recorded delivery time. The delivery delay for shipment $s_i$ is defined as:
\begin{equation}
    y_i = t_i^{\text{actual}} - t_i^{\text{planned}},
\end{equation}
where $y_i$ is measured in days. A shipment $s_i$ is classified as \textit{delayed} if $y_i \geq 1$ day, and \textit{on-time} otherwise. The one-day threshold reflects operational practice at the industrial partner, where sub-day deviations are typically absorbed within service-level buffers and do not trigger downstream actions.

The objective of this work is to learn a predictive mapping from shipment features $\mathbf{x}_i$ to delay duration, enabling both accurate point estimation of $y_i$ and calibrated prediction intervals that quantify the uncertainty in the delay estimate for future shipments. To account for heterogeneity in delivery patterns across source locations arising from regional, operational, and infrastructural differences, the proposed approach adopts a location-specific modeling strategy. In particular, separate models are trained for each source location $L_k \in \mathcal{L}$ for $k \in \{1, \dots, |\mathcal{L}|\}$, allowing the learning process to capture location-dependent delay dynamics. The proposed methodology is described in detail in Section~\ref{sec: method}.

\section{Methodology} \label{sec: method}

\begin{figure}[t!]
    \centering
    \includegraphics[width=0.95\textwidth]{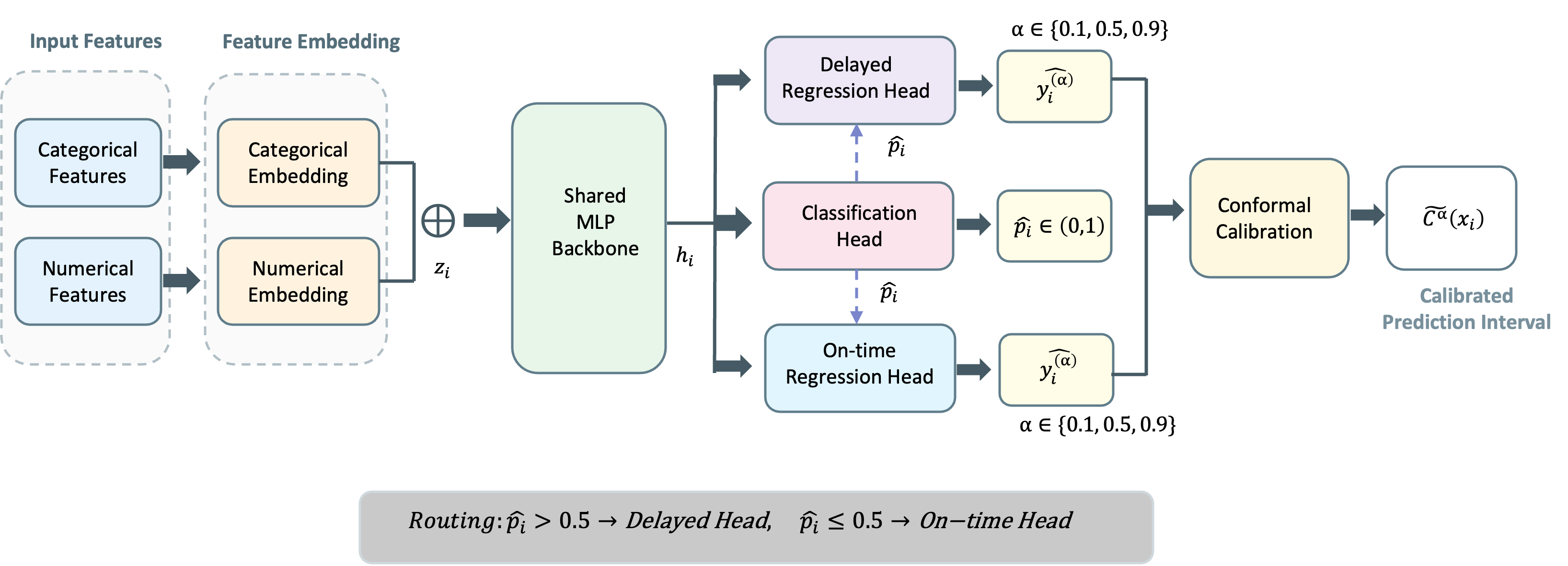}
    \caption{Overview of the multi-task learning model. Input features are processed through dedicated embedding layers (categorical and numerical), concatenated, and passed to a shared MLP backbone that produces a hidden representation $\mathbf{h}_i$. The backbone output feeds into three task-specific heads: a classification head for binary delay prediction, and dual regression heads for duration estimation. The classification probability $\hat{p}_i$ is concatenated with $\mathbf{h}_i$ and passed to both regression heads (dashed arrows). During training, samples are routed to regression heads based on ground truth labels; during inference, routing is determined by the classifier output ($\hat{p}_i > 0.5 \rightarrow$ delayed head). Quantile predictions are calibrated via conformal prediction to produce prediction intervals.}
    \label{fig:framework_overview}
\end{figure}

Figure~\ref{fig:framework_overview} illustrates the proposed model for delivery delay prediction with calibrated uncertainty. The model consists of four main components: (1) a shared embedding backbone with dedicated embedding layers for categorical and numerical features; (2) a classification head to predict the delay status, which acts as a routing mechanism to assign instances to the appropriate regression head; (3) dual quantile regression heads, which predict the delay duration for on-time and delayed instances separately; and (4) a conformal calibration step to adjust the prediction intervals.

During inference, to predict the delay for a shipment $s_i$, the model first encodes the feature vector $\mathbf{x}_i$ using the dedicated embedding layers. Next, the classifier predicts the delay status given the embedding vector $\mathbf{h}_i$. If the shipment is predicted to be delayed ($\hat{d}_i = 1$), it is routed to the delayed regression head; otherwise, it is routed to the on-time head. The model returns both a point estimate of the delay duration $\hat{y}_i^{(0.5)}$ and a calibrated prediction interval $\tilde{C}^\alpha(\mathbf{x}_i)=[\tilde{y}_i^{\text{low}}, \tilde{y}_i^{\text{high}}]$, where $\alpha \in (0, 1)$ is the user-specified miscoverage rate (e.g., $\alpha=0.2$ for 80\% coverage).

The proposed model is trained end-to-end via empirical risk minimization using training data $\mathcal{D}_{\text{train}}=\{(\mathbf{x}_i, y_i)\}_{i=1}^{N}$. The following sections detail the model architecture and training process.

\subsection{Feature Embedding}

Let $\mathbf{x}_i = [\mathbf{x}_i^{\text{cat}}, \mathbf{x}_i^{\text{num}}]$ denote the feature vector of shipment $s_i$, where $\mathbf{x}_i^{\text{cat}}=[x_{i,1}^{\text{cat}}, \dots, x_{i,K}^{\text{cat}}]$ represents $K$ categorical features and $\mathbf{x}_i^{\text{num}}=[x_{i,1}^{\text{num}}, \dots, x_{i,M}^{\text{num}}]$ represents $M$ numerical features. The embedding process for these features is as follows:

\paragraph{Categorical Embedding.} Denote $x^{\text{cat}}_{.,k}$ as a categorical feature with cardinality $C_k$, for $k \in \{1, \dots, K\}$. Each unique value of $x_{.,k}^{\text{cat}}$ is mapped to a $d^{\text{cat}}_{k}$-dimensional learnable vector via an embedding lookup:
\begin{equation}
    \text{Emb}(x_{i,k}^{\text{cat}}) = \mathbf{W}_k[x_{i,k}^{\text{cat}}] \in \mathbb{R}^{d_k^{\text{cat}}},
\end{equation}
where $\mathbf{W}_k \in \mathbb{R}^{C_k \times d_k^{\text{cat}}}$ is a learnable embedding matrix and $\mathbf{W}_k[x_{i,k}^{\text{cat}}]$ denotes the row corresponding to the category index. The embedding dimension $d^{\text{cat}}_{k}$ is specified as:
\begin{equation}
    d^{\text{cat}}_{k} = \min(d_{\max}^{\text{cat}}, \lfloor \log_2(C_k) + 1 \rfloor),
\end{equation}
where $d^{\text{cat}}_{\max}=50$ is set as the maximum embedding dimension. The logarithmic scaling ensures that embedding dimensions grow sublinearly with cardinality, providing sufficient representational capacity while preventing overparameterization. This approach enables the model to learn dense vector representations of categorical variables, capturing complex relationships by mapping similar categories close to each other in the embedding space~\cite{guoEntityEmbeddingsCategorical2016}.

\paragraph{Numerical Embedding.} Denote $x_{.,m}^{\text{num}}$ as a numerical feature, for $m \in \{1, \dots, M\}$. The model utilizes the Periodic Linear (PLR) embedding approach introduced by Gorishniy et al.~\cite{gorishniyEmbeddingsNumericalFeatures2023a} to map each numerical feature to a $d_m^{\text{num}}$-dimensional vector as follows:
\begin{equation}
    \text{Emb}(x_{i,m}^{\text{num}}) = \text{ReLU}\left(\mathbf{W}_m\left[\sin(2\pi \mathbf{c}_m x_{i,m}^{\text{num}}); \cos(2\pi \mathbf{c}_m x_{i,m}^{\text{num}})\right] + \mathbf{b}_m\right),
\end{equation}
where $\mathbf{c}_m = [c_1, \dots, c_\ell] \in \mathbb{R}^{\ell}$ are learnable frequency coefficients, $\mathbf{W}_m \in \mathbb{R}^{d_m^{\text{num}} \times 2\ell}$ is a learnable weight 
matrix, and $\mathbf{b}_m \in \mathbb{R}^{d_m^{\text{num}}}$ is a learnable bias vector.

The embedded categorical and numerical features are concatenated to form $\mathbf{z}_i \in \mathbb{R}^{d_z}$, where $d_z = \sum_{k=1}^{K} d_k^{\text{cat}} + \sum_{m=1}^{M} d_m^{\text{num}}$. The final representation is then obtained by passing $\mathbf{z}_i$ through a shared MLP backbone:
\begin{equation}
    \mathbf{h}_i = \text{MLP}(\mathbf{z}_i) \in \mathbb{R}^{d_h},
\end{equation}
where $\text{MLP}(\cdot)$ consists of sequential fully connected layers with ReLU activations and dropout regularization, and $d_h$ is the dimension of the hidden representation. The number of layers and $d_h$ are determined through hyperparameter tuning.

\subsection{Classification Head}

This component predicts whether a shipment is expected to be delayed. Let $d_i \in \{0,1\}$ denote the delay indicator derived from the observed delay duration, where $d_i = 1$ indicates a delayed shipment (i.e., $y_i \geq 1$ day). The classifier is a shallow MLP that takes as input the embedded feature representation vector $\mathbf{h}_i$ and predicts the delay status as follows:
\begin{equation} \label{eq:class_prob}
    \hat{p}_i = f_c(\mathbf{h}_i) = \sigma(\mathbf{w}_c^{T}\mathbf{h}_i + b_c),
\end{equation}
where $\mathbf{w}_c \in \mathbb{R}^{d_h}$ and $b_c \in \mathbb{R}$ are learnable parameters, and $\sigma(z) = 1/(1+e^{-z})$ is the sigmoid function. The predicted binary delay indicator is then obtained as:
\begin{equation}
    \hat{d}_i = \mathbb{I}\{\hat{p}_i > 0.5\}.
\end{equation}

The classifier acts as a routing mechanism by separating the dominant on-time regime from the delayed, allowing subsequent regression models to specialize and improve performance on delayed instances. To balance precision and recall, the SigmoidF1 loss function~\cite{bénédict2022sigmoidf1smoothf1score}, a smooth and differentiable surrogate for the F1-score, is used for training the classification head:

\begin{align}
    \mathcal{L}_{c} &= 1 - \widetilde{F1}, \\
    \widetilde{F1} &= \frac{2\tilde{tp}}{2\tilde{tp} + \tilde{fp} + \tilde{fn}}.
\end{align}

The soft confusion matrix components are calculated as follows:
\begin{align}
    \tilde{tp} &= \sum_{i=1}^{|\mathcal{B}|} \hat{p}_i \cdot d_i, \\
    \tilde{fp} &= \sum_{i=1}^{|\mathcal{B}|} \hat{p}_i \cdot (1 - d_i), \\
    \tilde{fn} &= \sum_{i=1}^{|\mathcal{B}|} (1 - \hat{p}_i) \cdot d_i,
\end{align}
where $\mathcal{B}$ denotes a mini-batch of training samples, $|\mathcal{B}|$ is the batch size, and $\hat{p}_i$ is the predicted delay probability from Eq.~\ref{eq:class_prob}, which serves as a smooth relaxation of the binary prediction.

This smooth approximation enables gradient-based optimization and is particularly effective for imbalanced classification as it jointly balances precision and recall. Note that unlike decomposable losses, SigmoidF1 is computed at the batch level since the F1 score requires aggregating confusion matrix statistics within each mini-batch.

\subsection{Quantile Regression Heads}

Given the classifier's output, the model estimates the delay duration using two specialized quantile regression heads. For a user-specified miscoverage level $\alpha \in (0, 1)$, the prediction interval is defined as $\hat{C}^{\alpha}(\mathbf{x}_i)=[\hat{y}_i^{(\alpha/2)}, \hat{y}_i^{(1-\alpha/2)}]$, with $\hat{y}_i^{(\alpha)}$ defined as:
\begin{equation}
    \hat{y}^{(\alpha)}_i =
    \begin{cases}
        f_r^{\text{delayed}}(\mathbf{r}_i;\, \alpha) & \text{if } \hat{d}_i = 1, \\
        f_r^{\text{ontime}}(\mathbf{r}_i;\, \alpha)  & \text{if } \hat{d}_i = 0,
    \end{cases}
\end{equation}
where $\mathbf{r}_i = [\mathbf{h}_i, \hat{p}_i]$ is the regression input formed by concatenating the hidden representation with the classification probability. This probability conditioning allows each regression head to leverage the classifier's confidence when making predictions. Each regression head is a shallow MLP with ReLU activations. The point estimate of delay duration is obtained as the median prediction $\hat{y}_i^{(0.5)}$.

Both regression heads are trained using the pinball loss (i.e., quantile loss)~\cite{Koenker_2005}. For a given quantile level $\alpha$ and prediction error $u = y_i - \hat{y}_i^{(\alpha)}$, the pinball loss is defined as:
\begin{equation}
    \rho_{\alpha}(u) = 
    \begin{cases} 
        \alpha \cdot u & \text{if } u \geq 0, \\ 
        (\alpha - 1) \cdot u & \text{if } u < 0.
    \end{cases}
\end{equation}

During training, samples are routed to regression heads based on ground truth delay labels ($d_i$). In this work, we target 80\% prediction intervals ($\alpha = 0.2$), corresponding to quantile levels $\{0.1, 0.5, 0.9\}$ for the lower bound, median, and upper bound respectively. The total regression loss is:


\begin{equation}
    \mathcal{L}_r = \frac{1}{3|\mathcal{B}|} \sum_{i=1}^{|\mathcal{B}|} \sum_{\alpha \in \{0.1, 0.5, 0.9\}} \left[ d_i \cdot \rho_{\alpha}(y_i - f_r^{\text{delayed}}(\mathbf{r}_i; \alpha)) + (1-d_i) \cdot \rho_{\alpha}(y_i - f_r^{\text{ontime}}(\mathbf{r}_i; \alpha)) \right]
\end{equation}
where $|\mathcal{B}|$ is the number of training instances in the mini-batch, and the $d_i$ and $(1-d_i)$ terms route each sample to the appropriate head.

Training is performed in two stages. First, the total loss $\mathcal{L}_t = \mathcal{L}_c + \mathcal{L}_r$ is minimized using mini-batch stochastic gradient descent over the training data. The classification loss is monitored and the first stage is stopped when no significant improvement is observed, or the maximum number of epochs is reached. In the second stage, all layers are frozen except the regression heads and training continues to minimize the regression loss, allowing each head to refine its quantile estimates without disrupting the shared representations and classification routing learned in the first stage.

\subsection{Conformal Calibration} 

The resulting intervals obtained from the quantile regression models may have high miscoverage rate due to the lack of proper training, or distribution shift in the data. To enhance reliability of predictions, the proposed model applies conformalized quantile regression (CQR) \cite{romanoConformalizedQuantileRegression2019} using a hold-out calibration set $\mathcal{D}_{\text{calib}} = \{(\mathbf{x}_i, y_i)\}_{i=1}^{N_{\text{calib}}}$.

The dual-head architecture requires separate calibration for each regression head. The calibration set is partitioned into two subsets based on ground truth labels: $\mathcal{D}_{\text{calib}}^{\text{delayed}} = \{i \in \mathcal{D}_{\text{calib}} : d_i = 1\}$ and $\mathcal{D}_{\text{calib}}^{\text{ontime}} = \{i \in \mathcal{D}_{\text{calib}} : d_i = 0\}$.

For each head, given the initial prediction interval $\hat{C}^{\alpha}(\mathbf{x}_i) = [\hat{y}_i^{(\alpha/2)}, \hat{y}_i^{(1-\alpha/2)}]$, conformity scores are calculated over the corresponding calibration subset:
\begin{equation}
    E_i^{\text{delayed}} = \max\left\{\hat{y}_i^{(\alpha/2), \text{delayed}} - y_i, \; y_i - \hat{y}_i^{(1-\alpha/2), \text{delayed}}\right\}, \quad i \in \mathcal{D}_{\text{calib}}^{\text{delayed}}
\end{equation}
with $E_i^{\text{ontime}}$ defined analogously for $i \in \mathcal{D}_{\text{calib}}^{\text{ontime}}$. The conformity score measures interval violation: $E_i > 0$ when the true value falls outside the predicted interval, and $E_i \leq 0$ when covered.

The calibration threshold for target coverage $1 - \alpha$ is the $(1-\alpha)$-quantile of the conformity scores:
\begin{equation}
    \hat{q}_{1-\alpha}^{\text{delayed}} = Q_{1-\alpha}\left(\{E_i^{\text{delayed}}\}_{i \in \mathcal{D}_{\text{calib}}^{\text{delayed}}}\right), \quad \hat{q}_{1-\alpha}^{\text{ontime}} = Q_{1-\alpha}\left(\{E_i^{\text{ontime}}\}_{i \in \mathcal{D}_{\text{calib}}^{\text{ontime}}}\right)
\end{equation}
 where $Q_{1-\alpha}(\cdot)$ denotes the empirical quantile function.

At test time, the calibrated prediction interval is obtained by adjusting the quantile predictions with the appropriate correction term:
\begin{equation}
    \tilde{C}^{\alpha}(\mathbf{x}_i) = \left[\hat{y}_i^{(\alpha/2)} - \hat{q}_{1-\alpha}, \; \hat{y}_i^{(1-\alpha/2)} + \hat{q}_{1-\alpha}\right]
\end{equation}
where the head selection (and corresponding $\hat{q}_{1-\alpha}$) is determined by the classifier: samples with $\hat{p}_i > 0.5$ use the delayed head and $\hat{q}_{1-\alpha}^{\text{delayed}}$, while samples with $\hat{p}_i \leq 0.5$ use the on-time head and $\hat{q}_{1-\alpha}^{\text{ontime}}$.

CQR provides finite-sample marginal coverage guarantees under exchangeability of calibration and test data \cite{romanoConformalizedQuantileRegression2019}. However, the temporal nature of logistics data violates this assumption. This paper validates coverage empirically on hold-out test data using chronological splits rather than claiming theoretical guarantees.

\section{Experimental Setup} \label{sec: exp_setup}
This section outlines the dataset, baseline, and the experimental setup for evaluating the proposed delivery delay prediction model.

\begin{figure}[t!]
   \centering
   \begin{subfigure}[b]{0.48\textwidth}
       \centering
       \includegraphics[width=\textwidth]{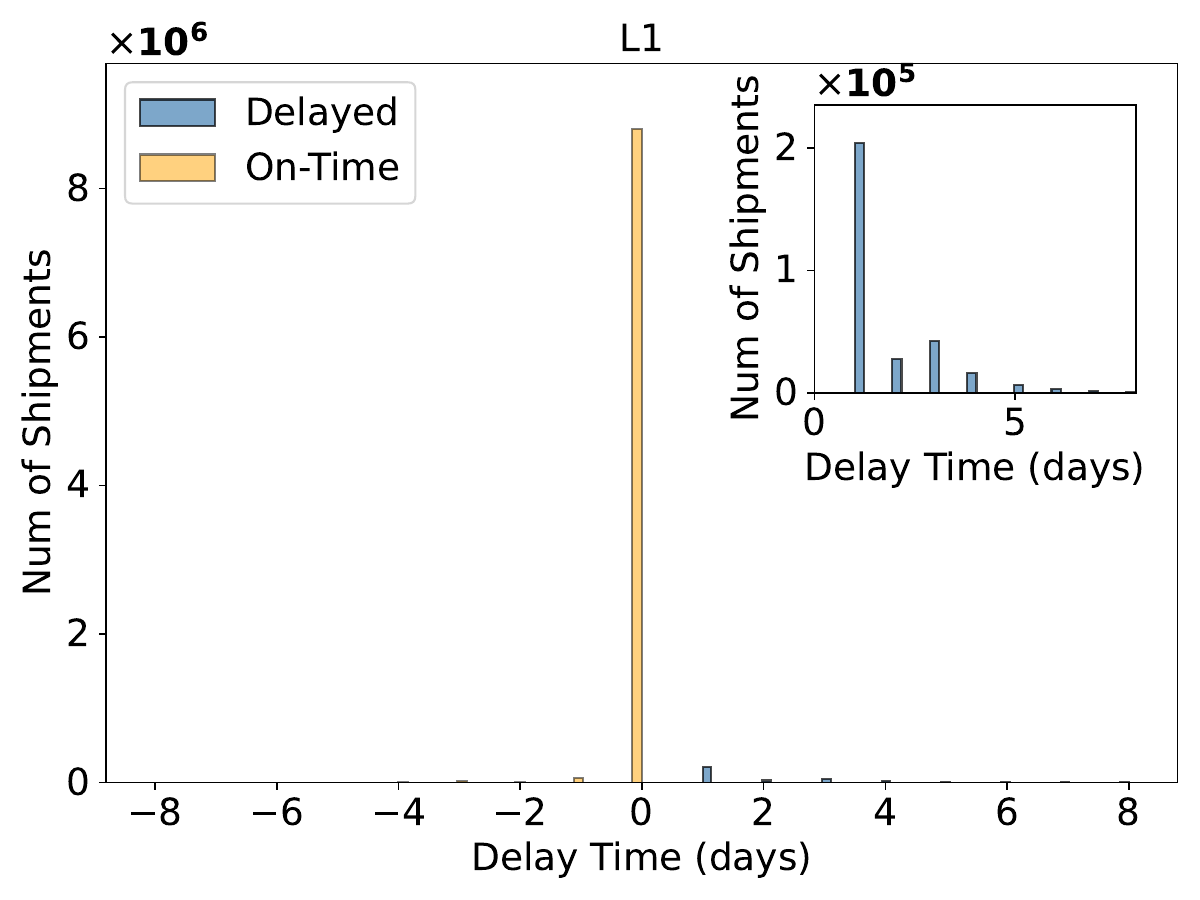}
       \caption{\centering L\textsubscript{1}: High volume}
       \label{fig:dist_a}
   \end{subfigure}
   \hspace{0.02\textwidth}
   \begin{subfigure}[b]{0.48\textwidth}
       \centering
       \includegraphics[width=\textwidth]{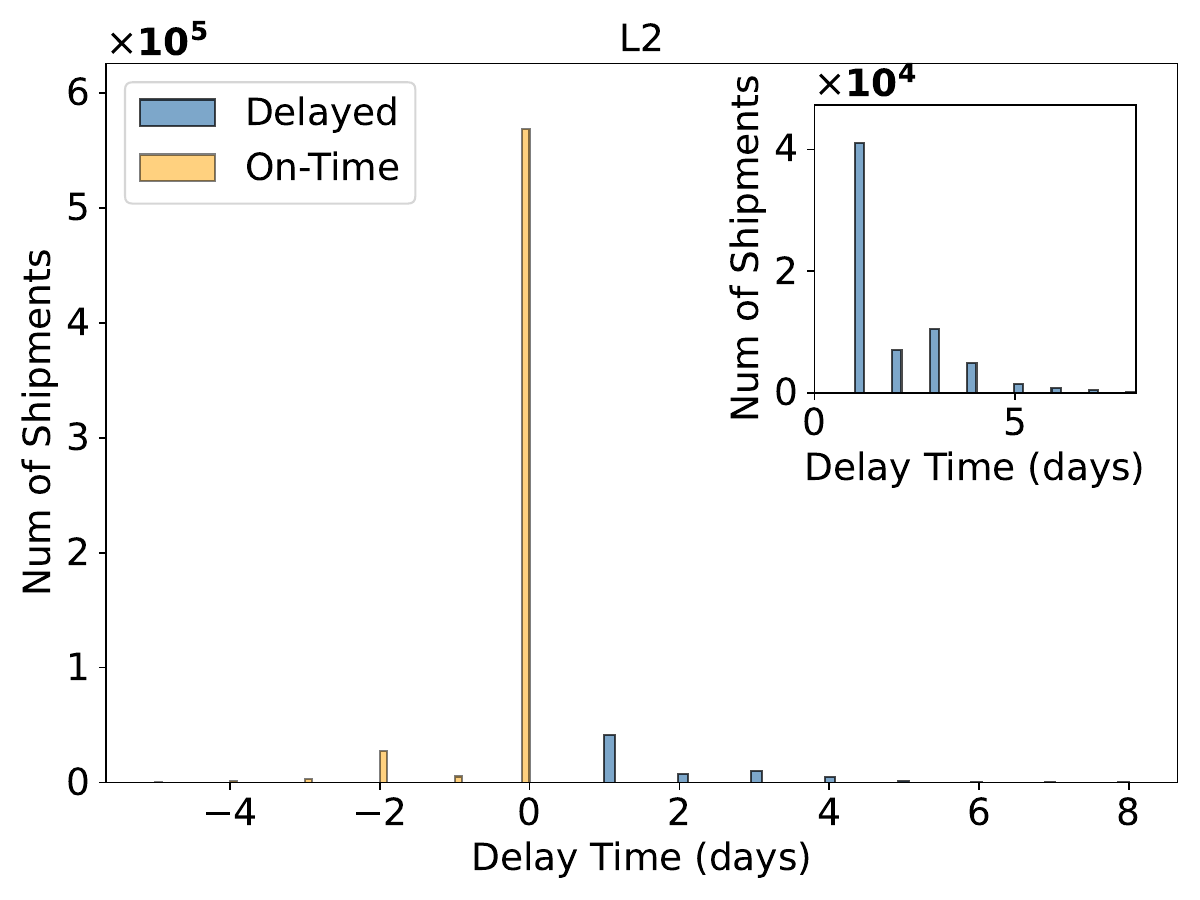}
       \caption{\centering L\textsubscript{2}: Moderate volume }
       \label{fig:dist_b}
   \end{subfigure}
   
   \vspace{0.05\textwidth}
   
   \begin{subfigure}[b]{0.48\textwidth}
       \centering
       \includegraphics[width=\textwidth]{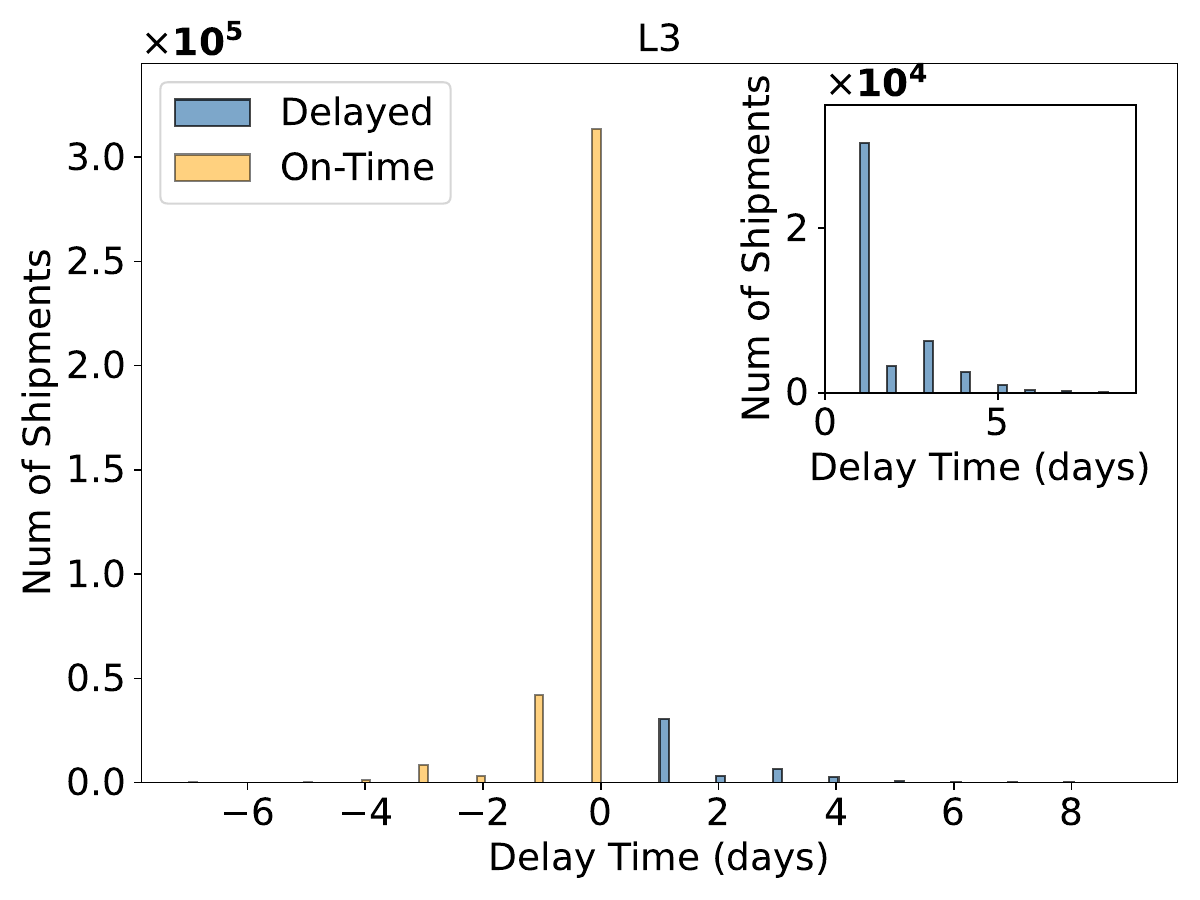}
       \caption{\centering L\textsubscript{3}: Moderate volume}
       \label{fig:dist_c}
   \end{subfigure}
   \hspace{0.02\textwidth}
   \begin{subfigure}[b]{0.48\textwidth}
       \centering
       \includegraphics[width=\textwidth]{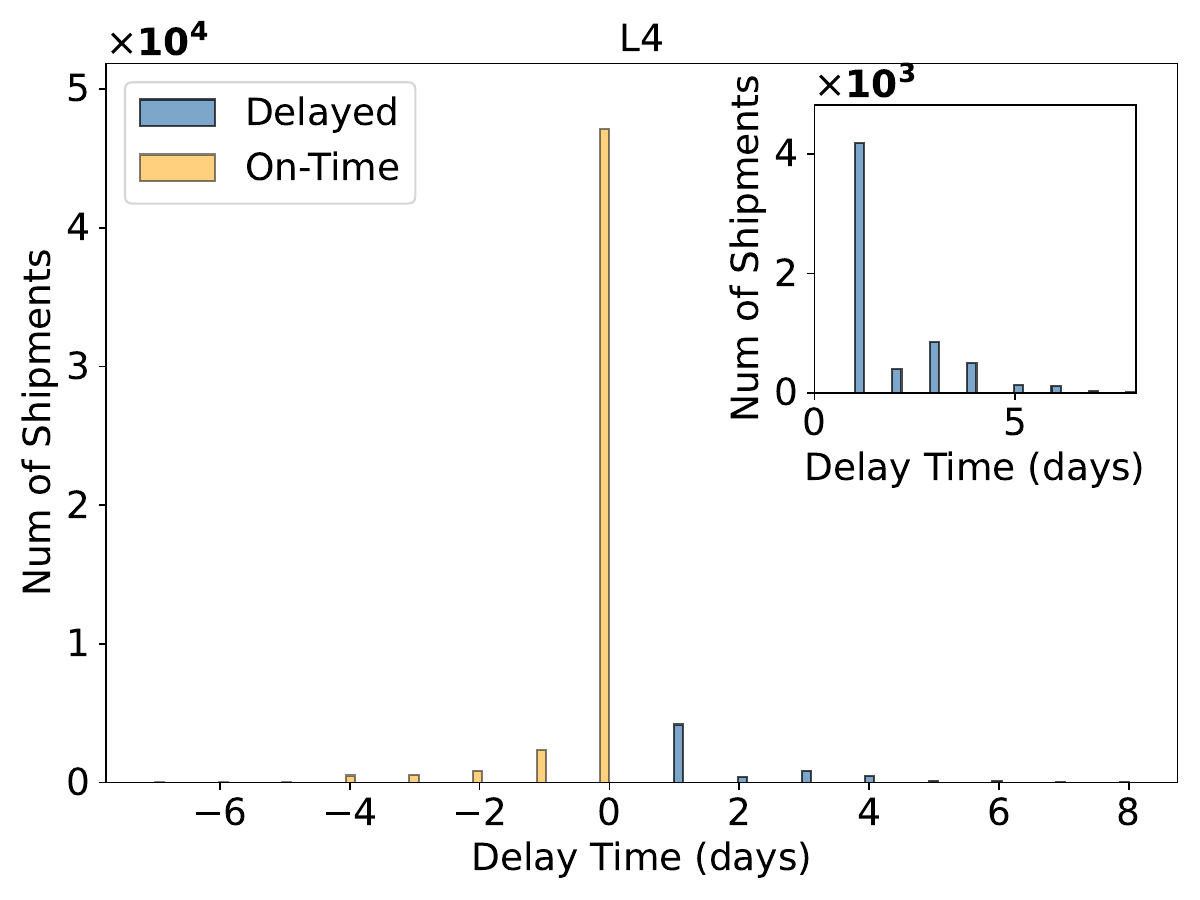}
       \caption{\centering L\textsubscript{4}: Low volume}
       \label{fig:dist_d}
   \end{subfigure}
    \caption{Delivery time distributions across locations showing distinct patterns of on-time (yellow) and delayed (blue) shipments. Main plots show the overall distribution, while insets focus on the delayed shipment patterns.}
   \label{fig:delivery_distributions}
\end{figure}

\subsection{Dataset Overview} \label{sec: Data}

\begin{figure}[t!]
   \centering
    \begin{minipage}{0.65\textwidth}
        \centering
        \includegraphics[width=\linewidth]{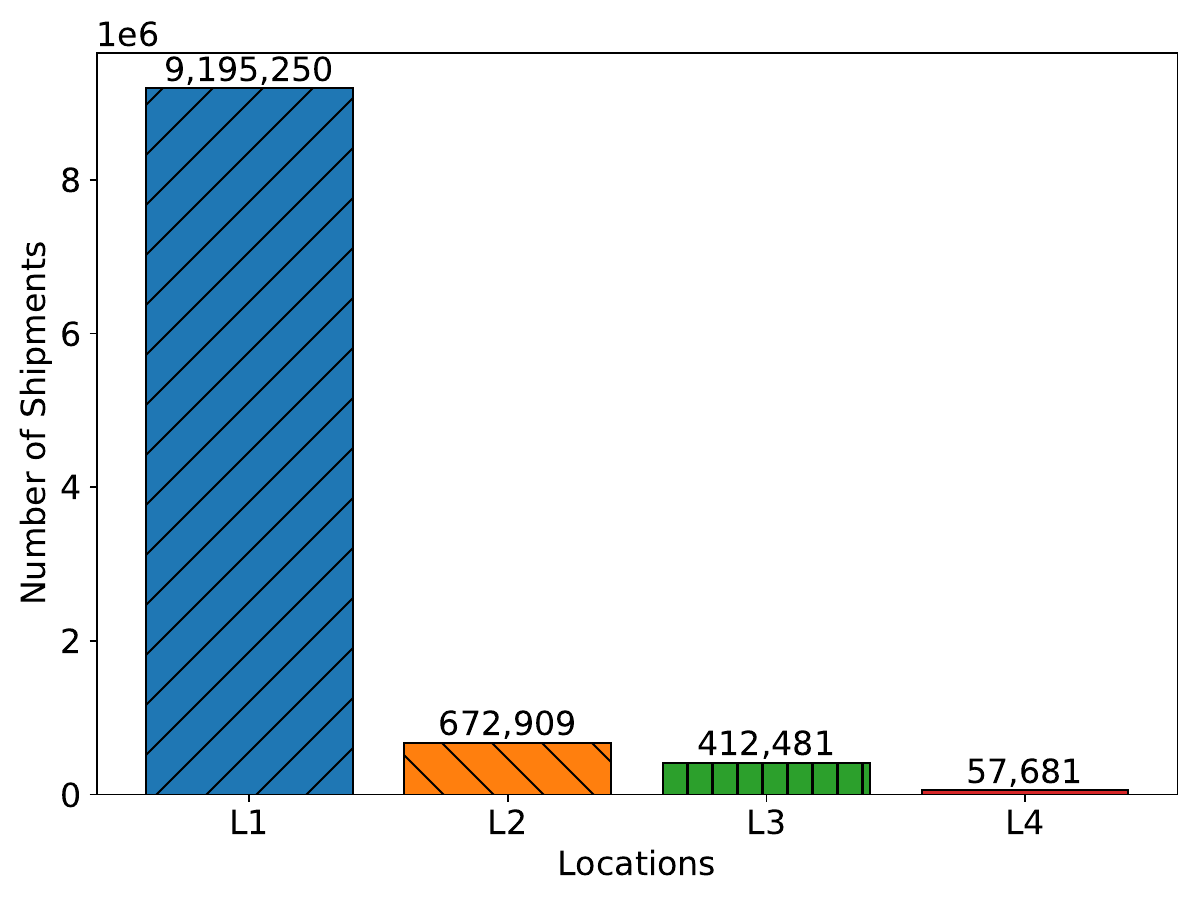}
        \caption{Distribution of shipment volumes across source locations.}
        \label{fig:shipment_volumes}
    \end{minipage}
\end{figure}

The dataset utilized in this study comprises over 10 million shipment records from a logistics provider, covering delivery operations conducted between September 2022 and January 2024. After preprocessing, the dataset includes four major source locations, referred to as $L_1, \dots, L_4$ (i.e., $|\mathcal{L}| = 4$). The number of shipments across different source locations is presented in Figure \ref{fig:shipment_volumes}, with $L_1$ being responsible for the majority of the shipments (over 9 million). Figure \ref{fig:delivery_distributions} presents the distribution of the delivery delays across different source locations. The majority of the shipments are delivered on time, with the delay percentage ranging from 3.28\% to 10.79\% across different source locations.

Each shipment in the dataset is associated with a combination of categorical and numerical features, as presented in Tables  \ref{tab:categorical_features} and \ref{tab:numerical_features}, respectively. The categorical features encompass location-related information (e.g., city, country) for both the origin and destination, hazardous materials, shipment type, and service type. The numerical features include the shipment's weight, volume, number of items, distance between the origin and destination, and the latitude and longitude of both the origin and destination. The dataset also includes temporal features, such as departure dates, planned arrival dates, and actual arrival dates. Delivery delays are calculated as the difference in days between the actual arrival date and the planned arrival date.

Due to the heterogeneity in operational scale and delay patterns across source locations, combined with the imbalanced distribution between delayed and on-time shipments, location-specific models are developed for each of the source locations $L_1-L_4$. Model architecture and the training process are similar across all locations, unless stated otherwise.

\begin{table}[b!]
\centering
\caption{Summary of Categorical Features}
\label{tab:categorical_features}
\resizebox{0.95\textwidth}{!}{%
\begin{tabular}{l r p{7cm}}
\toprule
\textbf{Feature} & \textbf{Cardinality} & \textbf{Description} \\
\midrule
From Location ID & 4 & Source location identifier \\
From City & 4 & City of shipment origin \\
From Postal Code & 4 & Postal code of origin \\
From Country System ID & 4 & Country code for origin \\
To Location ID & 198,019 & Destination location identifier \\
To City & 34,572 & Destination city name \\
To Postal Code & 52,236 & Destination postal code \\
To Country System ID & 29 & Country code for destination \\
Country Combination & 33 & Origin-destination country pair \\
Shipment Type ID & 37 & Classification of shipment type \\
Preferred Carrier ID & 81 & Identifier for the preferred shipping carrier \\
Dangerous Goods (Yes/No) & 2 & Indicator for hazardous materials (0/1) \\
\bottomrule
\end{tabular}
}
\end{table}

\begin{table}[b!]
\centering
\caption{Summary of Numerical Features}
\label{tab:numerical_features}
\resizebox{0.95\textwidth}{!}{%
\begin{tabular}{l l p{9cm}}
\toprule
\textbf{Feature} & \textbf{Unit} & \textbf{Description} \\
\midrule
Service Order Weight & Kilograms & Total weight of the shipment \\
Service Order Volume & Cubic meters & Total volume of the shipment \\
Total Number of Items & Count & Number of individual items in shipment \\
Destination City Latitude & Degrees & Destination city latitude coordinate \\
Destination City Longitude & Degrees & Destination city longitude coordinate \\
Distance & Kilometers & Direct distance between origin and destination \\
\bottomrule
\end{tabular}
}
\end{table}

\subsection{Baseline Methods}

The proposed model is compared with Gradient Boosting Decision Tree (GBDT) baselines, specifically eXtreme Gradient Boosting (XGBoost) \cite{chenXGBoostScalableTree2016} and Categorical Boosting (CatBoost) \cite{prokhorenkovaCatBoostUnbiasedBoosting2018}, under two different settings:
\begin{enumerate}
    \item \textbf{One-Step Prediction:} Under this setting, a single quantile regression model is trained to estimate the delay duration quantiles ($\alpha \in \{0.1, 0.5, 0.9\}$) for 80\% coverage. During the training, higher sample weights are employed for the delayed instances to account for the class imbalance.
    \item \textbf{Two-Step Prediction:} This setting involves a classification-then-regression approach where the two steps are trained separately, unlike the proposed model which trains end-to-end. First, a GBDT classifier is trained to predict the delay status. Next, two GBDT regression models are trained with pinball loss to predict the delay durations, one on delayed samples only and the other on on-time samples only. At inference time, shipments are routed to the corresponding regression head based on the classifier output. This approach mimics our dual-head architecture, but each regression model is trained independently on its respective subset, whereas our multitask model benefits from shared representations learned across all samples.
\end{enumerate}

\subsection{Experimental Setting}

All models are implemented in Python 3.9 using PyTorch 2.0 for neural network development and scikit-learn for traditional machine learning methods. Experiments are conducted on NVIDIA RTX 6000 GPUs with 16GB memory per task, supported by six CPU cores.

For each location, the data is first sorted based on the planned arrival times. Next, the data is divided into training, validation, calibration, and test sets with 0.7/0.1/0.1/0.1 ratios, respectively. This splitting strategy is used to avoid data leakage due to the temporal nature of the dataset. The validation data is used for hyperparameter tuning. After training the models, the calibration data is used for conformal calibration of the prediction intervals as explained in the Methodology section. Finally, all models are evaluated on the test set.

Deep learning models employ the AdamW optimizer with learning rate scheduling. The scheduler implements a linear warmup followed by a linear decay, which helps stabilize the early training process. Training proceeds in mini-batches, with gradient clipping applied to prevent exploding gradients. Early stopping is implemented based on validation performance to prevent overfitting. Hyperparameter tuning is performed using Optuna for both the machine learning baseline methods and our multitask model~\cite{akiba2019optuna}. The range of parameters for each method is presented in Appendix \ref{sec: appendix}.

Table \ref{tab:metrics} summarizes the evaluation metrics. All metrics are computed separately for overall samples and delayed samples only, denoted as (O/D) in the results tables. Let $\mathcal{D}_{\text{test}}^{*}$ denote the evaluation set: for overall metrics (O), $\mathcal{D}_{\text{test}}^{*} = \mathcal{D}_{\text{test}}$ includes all test samples; for delayed metrics (D), $\mathcal{D}_{\text{test}}^{*} = \mathcal{D}_{\text{test}}^{\text{delayed}}$ includes only shipments with $d_{i} = 1$. The point predictions are evaluated using Mean Absolute Error (MAE), comparing the ground truth ($y_i$) with the median prediction from the quantile regression models (i.e., $\hat{y}_i^{(0.5)}$). Additionally, the probabilistic predictions are evaluated using Coverage, Average Interval Width (AIW), and Winkler Score (WS) prior and after the conformal calibration steps.

\begin{table}[t!]
\caption{Summary of Evaluation Metrics}
\label{tab:metrics}
\resizebox{\textwidth}{!}{%
\begin{tabular}{lp{7cm}p{7cm}}
\toprule
\textbf{Metric} & \textbf{Formula} & \textbf{Description} \\
\midrule
\multicolumn{3}{l}{\textit{Delay Duration Estimation}} \\
\midrule
MAE (O/D) & 
$\displaystyle \frac{1}{\lvert\mathcal{D}_{\text{test}}^{*}\rvert}\sum_{i \in \mathcal{D}_{\text{test}}^{*}} \lvert y_i - \hat{y}_i^{(0.5)} \rvert$ & 
Mean absolute error between actual delays and median predictions (days). \\[10pt]
Avg-QL (O/D) & 
$\displaystyle \frac{1}{3\lvert\mathcal{D}_{\text{test}}^{*}\rvert}\sum_{\alpha \in \{0.1,0.5,0.9\}} \sum_{i \in \mathcal{D}_{\text{test}}^{*}} \rho_{\alpha}(y_i - \hat{y}_i^{(\alpha)})$ & 
Average quantile loss across all three quantiles, where $\rho_{\alpha}(u) = \max(\alpha u, (\alpha-1)u)$. \\[10pt]
\midrule
\multicolumn{3}{l}{\textit{Uncertainty Quantification}} \\
\midrule
Coverage (O/D) & 
$\displaystyle \frac{1}{\lvert\mathcal{D}_{\text{test}}^{*}\rvert}\sum_{i \in \mathcal{D}_{\text{test}}^{*}} \mathbb{I}[\hat{y}_i^{(0.1)} \leq y_i \leq \hat{y}_i^{(0.9)}]$ & 
Proportion of true delays within prediction intervals. Target: 80\% ($\alpha=0.2$). \\[10pt]
AIW (O/D) & 
$\displaystyle \frac{1}{\lvert\mathcal{D}_{\text{test}}^{*}\rvert}\sum_{i \in \mathcal{D}_{\text{test}}^{*}} (\hat{y}_i^{(0.9)} - \hat{y}_i^{(0.1)})$ & 
Average interval width (days). \\[10pt]
WS (O/D) &
\begin{tabular}[t]{@{}l@{}}
$\displaystyle \frac{1}{\lvert\mathcal{D}_{\text{test}}^{*}\rvert}
\sum_{i \in \mathcal{D}_{\text{test}}^{*}} \text{WS}_i,$ \\[15pt]
$\displaystyle \text{WS}_i =
\begin{cases}
w_i + \frac{2}{\alpha}(\hat{y}_i^{(0.1)} - y_i) & y_i < \hat{y}_i^{(0.1)} \\
w_i + \frac{2}{\alpha}(y_i - \hat{y}_i^{(0.9)}) & y_i > \hat{y}_i^{(0.9)} \\
w_i & \text{otherwise}
\end{cases}$
\end{tabular}
&
Winkler Score, where $w_i = \hat{y}_i^{(0.9)} - \hat{y}_i^{(0.1)}$. Penalizes interval width plus miscoverage. \\[10pt]
\bottomrule
\end{tabular}
}
\end{table}

\newpage

\section{Results} \label{sec: results}

This section presents the numerical evaluation and comparison of the proposed deep learning model against baseline methods for delivery delay duration estimation. Performance is evaluated across four source locations ($L_1, L_2, L_3, L_4$) using the metrics defined in Section \ref{sec: exp_setup}. The models include: 1) single-step quantile regression XGBoost and CatBoost referred to as XGB-S1 and CatB-S1; 2) two-step classify-then-regression models referred to as XGB-S2 and CatB-S2; and 3) the proposed multitask deep learning model, referred to as DL (ours) in the tables.

Table \ref{tab:point_prediction} presents the MAE and average quantile loss (Avg-QL, also known as pinball loss) for the delayed instances and the overall dataset across $L_1$--$L_4$. The proposed DL model achieves the lowest MAE and Avg-QL for the delayed instances across all the locations, with MAE ranging from 0.67 to 0.91. In particular, the proposed DL model improves the MAE on the delayed instances by 41--64\% compared to the best single-step model, and by 15--35\% compared to the best two-step tree-based model across the four locations.

The two-step tree-based models achieve slightly better performance on the delayed instances compared to the single-step models. However, their MAE on the overall instances is significantly higher, especially on the location with the highest number of shipments (i.e., $L_1$). In contrast, the proposed DL model achieves comparable MAE on the overall instances while achieving significantly lower error over the delayed shipments.

\begin{table*}[t]
\centering
\caption{Point prediction performance across locations. Best delayed performance per location highlighted in bold. Lower values indicate better performance. Ovr = overall instances, Del = delayed instances only.}
\label{tab:point_prediction}
\Large
\setlength{\tabcolsep}{2pt}
\renewcommand{\arraystretch}{1.05}
\resizebox{\linewidth}{!}{%
\begin{tabular}{l cccc cccc cccc cccc}
\toprule
\textbf{Model}
& \multicolumn{4}{c}{\textbf{L1}}
& \multicolumn{4}{c}{\textbf{L2}}
& \multicolumn{4}{c}{\textbf{L3}}
& \multicolumn{4}{c}{\textbf{L4}} \\
\cmidrule(lr){2-5} \cmidrule(lr){6-9} \cmidrule(lr){10-13} \cmidrule(lr){14-17}
& \multicolumn{2}{c}{Avg-QL $(\downarrow)$} & \multicolumn{2}{c}{MAE $(\downarrow)$}
& \multicolumn{2}{c}{Avg-QL $(\downarrow)$} & \multicolumn{2}{c}{MAE $(\downarrow)$}
& \multicolumn{2}{c}{Avg-QL $(\downarrow)$} & \multicolumn{2}{c}{MAE $(\downarrow)$}
& \multicolumn{2}{c}{Avg-QL $(\downarrow)$} & \multicolumn{2}{c}{MAE $(\downarrow)$} \\
\cmidrule(lr){2-3} \cmidrule(lr){4-5}
\cmidrule(lr){6-7} \cmidrule(lr){8-9}
\cmidrule(lr){10-11} \cmidrule(lr){12-13}
\cmidrule(lr){14-15} \cmidrule(lr){16-17}
& Ovr & Del & Ovr & Del
& Ovr & Del & Ovr & Del
& Ovr & Del & Ovr & Del
& Ovr & Del & Ovr & Del \\
\midrule
XGB-S1   & 0.05 & 0.52 & 0.10 & 1.43 & 0.13 & 0.58 & 0.33 & 1.48 & 0.17 & 0.59 & 0.41 & 1.85 & 0.16 & 0.64 & 0.32 & 1.70 \\
CatB-S1  & 0.05 & 0.76 & 0.11 & 1.72 & 0.14 & 0.59 & 0.41 & 1.59 & 0.18 & 0.62 & 0.41 & 1.86 & 0.16 & 0.79 & 0.30 & 1.86 \\
XGB-S2   & 0.11 & 0.46 & 0.23 & 1.02 & 0.20 & 0.52 & 0.40 & 1.16 & 0.28 & 0.40 & 0.60 & 0.90 & 0.20 & 0.67 & 0.41 & 1.40 \\
CatB-S2  & 0.10 & 0.47 & 0.21 & 1.00 & 0.17 & 0.55 & 0.37 & 1.24 & 0.32 & 0.33 & 0.70 & 0.81 & 0.19 & 0.65 & 0.34 & 1.40 \\
\midrule
DL (ours) & 0.05 & \textbf{0.39} & 0.12 & \textbf{0.84} & 0.14 & \textbf{0.39} & 0.37 & \textbf{0.85} & 0.19 & \textbf{0.30} & 0.45 & \textbf{0.67} & 0.17 & \textbf{0.42} & 0.35 & \textbf{0.91} \\
\bottomrule
\end{tabular}
}
\end{table*}

Tables \ref{tab:uncertainty_pre} and \ref{tab:uncertainty_post} present the uncertainty quantification performance evaluation for the baselines and the proposed DL model before and after conformal calibration, respectively. The metrics are calculated on the outputs of the quantile regression models without any calibration (Table~\ref{tab:uncertainty_pre}), and after performing the CQR approach (Table~\ref{tab:uncertainty_post}) for the target coverage of 80\%.

Considering the coverage rates of the models prior to the conformal calibration (Table~\ref{tab:uncertainty_pre}), most models achieve overall coverage above 70\%, except for the CatBoost-based models which exhibit substantially lower coverage at several locations. Comparing the results on the coverage over the delayed instances, the proposed DL model shows significantly higher coverage (ranges from 63.6\% to 69.8\%) compared to the other baselines. The best single-step model's coverage ranges from 19.7\% to 48.0\%. The best two-step models also achieve coverage rates of 21.7\% to 58.4\% across different locations, falling behind the proposed model. The conformal calibration improves the coverage of the models over both overall samples and the delayed instances, except for the single-step models. For the single-step models, due to the lack of any classification step, the calibration strategy does not differentiate between the delayed and on-time instances and as a result the conditional coverage does not improve significantly. However, in the proposed DL model the calibration method improves the coverage rate on both overall and the delayed instances. 

Note that due to the temporal nature of the data and the possibility of distribution shift, the exchangeability assumption required for conformal prediction's statistical guarantees does not hold for any of the models. Additionally, the two-step tree-based models may suffer from error propagation between the classification and regression stages. However, the empirical results demonstrate the efficacy of CQR in improving the calibration of the prediction intervals.

The AIW values of the calibrated prediction intervals (Table~\ref{tab:uncertainty_post}) range from 1.32 to 1.57 days for the proposed DL model on the delayed samples, and from 0.84 to 1.14 over all instances. While some of the baselines seem to provide sharper prediction intervals, they often exhibit a significant miscoverage rate. To fairly compare the trade-off between the AIW and the coverage rates, the paper considers Winkler Score (WS). As shown in Tables \ref{tab:uncertainty_pre} and \ref{tab:uncertainty_post}, the proposed DL model results in substantially lower WS for the prediction of the delayed instances over all locations. Considering the WS over all instances, the proposed model is slightly worse than the single-step models. However, it provides substantially more reliable intervals for the delayed instances, which are operationally important.

\begin{table*}[t]
\centering
\caption{Uncertainty quantification performance (pre-calibration). Target coverage is 80\% ($\alpha=0.2$). Best delayed performance per location highlighted in bold for Coverage and WS. For Coverage, higher is better; for AIW and WS, lower is better.}
\label{tab:uncertainty_pre}
\Large
\setlength{\tabcolsep}{2pt}
\renewcommand{\arraystretch}{1.05}
\resizebox{\linewidth}{!}{%
\begin{tabular}{l l ccc ccc ccc ccc}
\toprule
\textbf{Model} & \textbf{Type}
& \multicolumn{3}{c}{\textbf{L1}}
& \multicolumn{3}{c}{\textbf{L2}}
& \multicolumn{3}{c}{\textbf{L3}}
& \multicolumn{3}{c}{\textbf{L4}} \\
\cmidrule(lr){3-5} \cmidrule(lr){6-8} \cmidrule(lr){9-11} \cmidrule(lr){12-14}
& & Cov $\uparrow$ & AIW $\downarrow$ & WS $\downarrow$
& Cov $\uparrow$ & AIW $\downarrow$ & WS $\downarrow$
& Cov $\uparrow$ & AIW $\downarrow$ & WS $\downarrow$
& Cov $\uparrow$ & AIW $\downarrow$ & WS $\downarrow$ \\
\midrule
\multirow{2}{*}{XGB-S1}
& Overall & 0.843 & 0.381 & 0.926 & 0.769 & 0.650 & 2.135 & 0.823 & 0.663 & 3.128 & 0.700 & 0.438 & 3.131 \\
& Delayed & 0.435 & 1.442 & 8.411 & 0.422 & 1.406 & 9.914 & 0.455 & 1.346 & 8.453 & 0.197 & 0.862 & 10.768 \\
\cmidrule(lr){2-14}
\multirow{2}{*}{CatB-S1}
& Overall & 0.729 & 0.412 & 1.124 & 0.648 & 0.653 & 2.105 & 0.561 & 0.562 & 3.232 & 0.642 & 0.392 & 3.284 \\
& Delayed & 0.358 & 0.986 & 9.542 & 0.480 & 1.493 & 9.906 & 0.350 & 1.017 & 9.471 & 0.103 & 0.741 & 13.842 \\
\cmidrule(lr){2-14}
\multirow{2}{*}{XGB-S2}
& Overall & 0.864 & 0.184 & 2.255 & 0.769 & 0.438 & 3.983 & 0.647 & 0.719 & 5.481 & 0.768 & 0.153 & 4.095 \\
& Delayed & 0.460 & 0.865 & 8.807 & 0.532 & 1.100 & 9.762 & 0.584 & 1.494 & 7.592 & 0.217 & 0.381 & 13.213 \\
\cmidrule(lr){2-14}
\multirow{2}{*}{CatB-S2}
& Overall & 0.456 & 0.160 & 2.059 & 0.655 & 0.379 & 3.309 & 0.405 & 0.866 & 6.135 & 0.652 & 0.174 & 4.117 \\
& Delayed & 0.418 & 0.771 & 9.097 & 0.355 & 1.053 & 10.188 & 0.351 & 1.624 & 5.950 & 0.182 & 0.425 & 12.535 \\
\midrule
\multirow{2}{*}{DL (ours)}
& Overall & 0.816 & 0.448 & 1.324 & 0.798 & 0.728 & 2.964 & 0.742 & 0.886 & 4.218 & 0.772 & 0.598 & 4.042 \\
& Delayed & \textbf{0.682} & 0.894 & \textbf{6.742} & \textbf{0.664} & 1.008 & \textbf{7.284} & \textbf{0.698} & 1.164 & \textbf{5.412} & \textbf{0.636} & 0.774 & \textbf{6.986} \\
\bottomrule
\end{tabular}
}
\end{table*}

\begin{table*}[t]
\centering
\caption{Uncertainty quantification performance (post-calibration). Target coverage is 80\% ($\alpha=0.2$). Best delayed performance per location highlighted in bold for Coverage and WS. For Coverage, higher is better; for AIW and WS, lower is better.}
\label{tab:uncertainty_post}
\Large
\setlength{\tabcolsep}{2pt}
\renewcommand{\arraystretch}{1.05}
\resizebox{\linewidth}{!}{%
\begin{tabular}{l l ccc ccc ccc ccc}
\toprule
\textbf{Model} & \textbf{Type}
& \multicolumn{3}{c}{\textbf{L1}}
& \multicolumn{3}{c}{\textbf{L2}}
& \multicolumn{3}{c}{\textbf{L3}}
& \multicolumn{3}{c}{\textbf{L4}} \\
\cmidrule(lr){3-5} \cmidrule(lr){6-8} \cmidrule(lr){9-11} \cmidrule(lr){12-14}
& & Cov $\uparrow$ & AIW $\downarrow$ & WS $\downarrow$
& Cov $\uparrow$ & AIW $\downarrow$ & WS $\downarrow$
& Cov $\uparrow$ & AIW $\downarrow$ & WS $\downarrow$
& Cov $\uparrow$ & AIW $\downarrow$ & WS $\downarrow$ \\
\midrule
\multirow{2}{*}{XGB-S1}
& Overall & 0.846 & 0.384 & 0.924 & 0.847 & 0.659 & 2.132 & 0.826 & 0.664 & 3.126 & 0.774 & 0.440 & 3.129 \\
& Delayed & 0.438 & 1.448 & 8.396 & 0.426 & 1.418 & 9.869 & 0.457 & 1.348 & 8.450 & 0.199 & 0.865 & 10.762 \\
\cmidrule(lr){2-14}
\multirow{2}{*}{CatB-S1}
& Overall & 0.812 & 0.578 & 1.386 & 0.764 & 0.816 & 2.348 & 0.678 & 0.694 & 2.986 & 0.728 & 0.524 & 3.086 \\
& Delayed & 0.362 & 0.992 & 9.538 & 0.484 & 1.502 & 9.873 & 0.406 & 1.024 & 9.436 & 0.108 & 0.749 & 13.718 \\
\cmidrule(lr){2-14}
\multirow{2}{*}{XGB-S2}
& Overall & 0.870 & 0.186 & 2.248 & 0.776 & 0.443 & 3.972 & 0.654 & 0.726 & 5.468 & 0.772 & 0.155 & 4.086 \\
& Delayed & 0.466 & 0.871 & 8.794 & 0.538 & 1.108 & 9.746 & 0.591 & 1.504 & 7.574 & 0.221 & 0.385 & 13.194 \\
\cmidrule(lr){2-14}
\multirow{2}{*}{CatB-S2}
& Overall & 0.692 & 1.124 & 4.824 & 0.742 & 0.512 & 3.126 & 0.687 & 1.314 & 5.050 & 0.681 & 0.196 & 4.024 \\
& Delayed & 0.596 & 1.886 & 6.982 & 0.586 & 1.384 & 8.642 & 0.580 & 2.136 & 7.143 & 0.228 & 0.468 & 12.186 \\
\midrule
\multirow{2}{*}{DL (ours)}
& Overall & 0.852 & 0.839 & 1.089 & 0.854 & 1.016 & 2.524 & 0.828 & 1.082 & 3.742 & 0.822 & 1.142 & 3.486 \\
& Delayed & \textbf{0.726} & 1.324 & \textbf{5.864} & \textbf{0.718} & 1.482 & \textbf{6.642} & \textbf{0.742} & 1.486 & \textbf{4.982} & \textbf{0.686} & 1.568 & \textbf{6.124} \\
\bottomrule
\end{tabular}
}
\end{table*}

\newpage 

\section{Conclusion and Future Research}

This paper presented a multitask deep learning model for delivery delay duration estimation using a classification-then-regression strategy. The proposed model addresses a key limitation of existing approaches: standard regression models trained on highly imbalanced logistics data tend to optimize for the majority on-time class, producing poor predictions for the operationally critical delayed shipments. The proposed model utilizes dedicated numerical and categorical embedding layers to process high-dimensional tabular features. By employing dual regression heads with a routing mechanism based on classification confidence, the model enables each regression head to specialize on its target distribution while benefiting from shared feature representations learned across all training samples.

The proposed model was evaluated on a large-scale dataset from an industrial partner containing over 10 million shipment records across four major source locations, each exhibiting distinct regional characteristics and delay rates ranging from 3\% to 11\%. Experimental results demonstrated that the proposed multitask approach achieves mean absolute errors of 0.67--0.91 days for delayed shipments, representing improvements of 41\%--64\% over single-step tree-based baselines and 15\%--35\% over sequential two-step approaches. The model also produces significantly better-calibrated prediction intervals for delayed shipments, achieving 64\%--70\% coverage before calibration compared to 20\%--48\% for the best single-step tree-based baselines. After conformal calibration, the proposed model reaches the nominal 80\% coverage on overall shipments while consistently maintaining the best Winkler Scores for delayed shipments across all locations.

From a practical standpoint, the model enables logistics planners to identify shipments at risk of significant delay with greater accuracy. The calibrated prediction intervals provide operationally useful uncertainty estimates that support decision-making. As noted by our industry partner, the ability to identify shipments that are likely to be substantially late rather than borderline cases aligns well with core operational priorities.

Several directions remain for future work. Transfer learning approaches could improve performance in data-sparse locations by leveraging representations learned from high-volume origins. Incorporating temporal dynamics through online learning could enable the model to adapt to evolving delivery patterns and seasonal variations. Finally, extending the uncertainty quantification to provide conditional coverage guarantees for specific subgroups could further enhance operational utility.

\section*{Acknowledgments}
This research was partly supported by the NSF AI Institute for Advances in Optimization (Award
2112533).

\bibliographystyle{ieeetr}
\bibliography{references}
 
\newpage 

\appendix

\section{Appendix} \label{sec: appendix}

This appendix summarizes the hyperparameter search spaces used for tuning all models. Hyperparameter optimization was performed using Optuna with the Tree-structured Parzen Estimator (TPE) sampler, with 100 trials per model and location.

\subsection{GBDT Baselines}

Tables \ref{tab:xgb_params} and \ref{tab:catb_params} present the hyperparameter search ranges for the XGBoost and CatBoost models used in both one-step and two-step baseline configurations. The ranges were chosen based on prior work and preliminary experiments to ensure robust search spaces.

\begin{table}[h!]
\centering
\caption{Hyperparameter Search Space for XGBoost}
\label{tab:xgb_params}
\begin{tabular}{l l l}
\toprule
\textbf{Parameter} & \textbf{Range} & \textbf{Scale} \\
\midrule
n\_estimators & [200, 2000] & linear \\
max\_depth & [3, 12] & linear \\
learning\_rate & [1e-3, 0.3] & log \\
subsample & [0.6, 1.0] & linear \\
colsample\_bytree & [0.6, 1.0] & linear \\
gamma & [0, 5] & linear \\
reg\_alpha & [0.0, 20.0] & linear \\
reg\_lambda & [0.0, 50.0] & linear \\
min\_child\_weight & [1, 10] & linear \\
\bottomrule
\end{tabular}
\end{table}

\begin{table}[h!]
\centering
\caption{Hyperparameter Search Space for CatBoost}
\label{tab:catb_params}
\begin{tabular}{l l l}
\toprule
\textbf{Parameter} & \textbf{Range} & \textbf{Scale} \\
\midrule
iterations & [200, 1500] & linear \\
depth & [4, 12] & linear \\
learning\_rate & [1e-3, 0.3] & log \\
l2\_leaf\_reg & [1, 30] & linear \\
border\_count & [32, 255] & linear \\
bagging\_temperature & [0, 10] & linear \\
random\_strength & [0.0, 10.0] & linear \\
min\_child\_samples & [5, 100] & linear \\
\bottomrule
\end{tabular}
\end{table}

\subsection{Multitask Deep Learning Model}

Table \ref{tab:mtl_dl_params} presents the hyperparameter search space for the proposed multitask deep learning model. The architecture consists of a shared MLP backbone with dedicated embedding layers for categorical and numerical features, followed by task-specific heads.

\begin{table}[h!]
\centering
\caption{Hyperparameter search space for the multitask deep learning model.}
\label{tab:mtl_dl_params}
\begin{tabular}{ll}
\toprule
\textbf{Parameter} & \textbf{Range / Values} \\
\midrule
n\_blocks & [2, 12] \\
d\_hidden & \{128, 192, 256, 320, 384\} \\
dropout & [0.0, 0.5] \\
weight\_decay & [1e-6, 3e-3] \\
gradient\_clip\_norm & \{None, 1.0, 5.0\} \\
batch\_size & \{256, 512, 1024, 2048\} \\
learning\_rate & [1e-4, 1e-2] (log scale) \\
\bottomrule
\end{tabular}
\end{table}

For the embedding layers, categorical embedding dimensions are computed as $d_k^{\text{cat}} = \min(d_{\max}^{\text{cat}}, \lfloor \log_2(C_k) + 1 \rfloor)$, where $d_{\max}^{\text{cat}} = 50$ and $C_k$ denotes the cardinality of categorical feature $k$. Numerical features are embedded using the Periodic Linear (PLR) embedding with embedding dimension 24. The base learning rate was tuned using Optuna, and the training utilized a linear warmup followed by a linear decay schedule.

\end{document}